\def\year{2021}\relax
\documentclass[letterpaper]{article} 
\usepackage{aaai21}  
\usepackage{times}  
\usepackage{helvet} 
\usepackage{courier}  
\usepackage[hyphens]{url}  
\usepackage{graphicx} 
\usepackage{natbib}  
\usepackage{amsmath}
\usepackage{multirow}
\usepackage{subcaption}
\usepackage{caption} 
\usepackage{ulem}

\title{Distribution Matching for Rationalization}
\author{
    Yongfeng Huang\textsuperscript{\rm 2}\footnote{Work done during an internship at Recurrent AI}, Yujun Chen\textsuperscript{\rm 1}, Yulun Du\textsuperscript{\rm 1}, Zhilin Yang\textsuperscript{\rm 1}
    \\
}
\affiliations{
    \textsuperscript{\rm 1}Recurrent AI, Beijing \\
    \textsuperscript{\rm 2}Tsinghua University, Beijing\\

    huangyf17@tsinghua.org.cn, chenyujun@rcrai.com, duyulun@rcrai.com, kimi\_yang@rcrai.com\\ 

}

\begin{document}

\maketitle

\begin{abstract}
The task of rationalization aims to extract pieces of input text as rationales to justify neural network predictions on text classification tasks. By definition, rationales represent key text pieces used for prediction and thus should have similar classification feature distribution compared to the original input text. However, previous methods mainly focused on maximizing the mutual information between rationales and labels while neglecting the relationship between rationales and input text. To address this issue, we propose a novel rationalization method that matches the distributions of rationales and input text in both the feature space and output space. Empirically, the proposed distribution matching approach consistently outperforms previous methods by a large margin. Our data and code are available\footnote{https://github.com/kochsnow/distribution-matching-rationality}.
\end{abstract}

\section{Introduction}
In many real-world NLP applications, interpretability is an important objective for model development because it is crucial for human users to understand, verify, and trust the machine predictions. Among other possibilities, rationalization is a learning paradigm that extracts key text pieces as rationales to justify and interpret model predictions \cite{lei2016rationalizing}. Specifically, \citet{lei2016rationalizing} uses a generator to selectively extract rationales from the original input, and a classifier is applied on the rationales to predict the classification labels. This can be viewed as a cooperative game between the generator and the classifier to maximize the mutual information between the rationales and the labels \cite{Chen2018LearningTE}. In other words, this is based on the desideratum that the extracted rationales are predictive of the classification labels. Different variants of rationalization methods have been proposed under this framework, which additionally consider other desiderata such as the dependency between labels, rationales, and the complement of rationales \cite{chang2019game,chang2020invariant,yu2019rethinking}.

In this work, we argue that it is crucial to incorporate the following desideratum into modeling---the rationales and the original full input text should have similar feature and output distributions when the same classifier is applied. By definition, rationales represent key text pieces that are actually used for predicting the labels. The definition has a two-folded implication. First, the rationales should have a similar feature distribution to the input text because intermediate feature representations directly reflect how the model processes natural language. Second, since ultimately the probability outputs are used for classification, the rationales should have a similar output distribution to the input text. For example, consider a review ``this is a great movie''. A well-trained sentiment classifier mainly uses the rationale ``great movie'' for prediction. When the classifier is applied, ``great movie'' and ``this is a great movie'' should have similar feature and output distributions.

However, the aforementioned prior has not been effectively leveraged in previous approaches. As a solution, we propose a novel distribution matching approach for rationalization. In the feature space, we impose a regularization term that minimizes the central moment discrepancy (CMD) \cite{Zellinger2017CentralMD} between the full input features and the rationale features. In the output space, a teacher-student distillation loss \cite{hinton2015distilling} is employed to minimize the cross entropy loss between the full input predictions and the rationale predictions. 
Our approach is a plug-and-play improvement that is applicable to different rationalization variants.

We evaluate the proposed distribution matching approach on widely-used rationalization benchmarks---the beer review dataset ~\cite{McAuley2012LearningAA} and hotel review dataset ~\cite{Bao2018DerivingMA}. We use the game-theoretic class-dependent model \cite{chang2019game} as our base model. Empirical results show that distribution matching substantially improves over the baseline and outperforms all considered previous methods. It is also observed that both feature space matching and output space matching contribute to the overall performance.

To summarize, our work makes the following four contributions: 
\begin{itemize}
    \item We analyze the rationalization framework and uncover the issue that existing approaches neglect the relationship between rationales and input text. 
    \item We propose to impose an inductive bias that rationales should have similar feature and output distributions with input text so as to improve the faithfulness of rationalization. 
    \item We achieved state-of-the-art results with substantial gains on multiple settings. 
\end{itemize}

\section{Related Work}
\subsection{Interpretability}
There are multiple lines of research in the area of learning interpretable models for NLP tasks. Roughly, there are three categories---post-hoc explanation methods, extractive retionalization methods, and the self-explaining model-based approach.

\paragraph{Extractive Rationalization.}
Extractive rationalization selects pieces of text from the input to form rationales that justify the prediction. Multiple variants were proposed to improve over the original framework \cite{lei2016rationalizing}. \citet{chang2019game} introduced a game-theoretic framework where the rationale generator is dependent on the class labels. \citet{yu2019rethinking} employed a similar idea and additionally employed constraints on the complement of rationales. Other approaches were based on the information bottleneck \cite{paranjape2020information}, latent variable models \cite{Bastings2019InterpretableNP}, and  learning environment-invariant representations \cite{chang2020invariant}.  However, none of these previous methods consider the relationship between rationales and the full input in terms of feature distribution. 

Recently, a new benchmark \cite{deyoung2019eraser} was introduced with labeled rationales available for training. This enables directly finetuning pretrained models \cite{DBLP:journals/corr/abs-1802-05365,Radford2018ImprovingLU,Devlin2019BERTPO,Yang2019XLNetGA} to predict the rationales in an end-to-end fashion. However, in this work, we focus on the conventional unsupervised learning setting because rationale groundtruth is not available for most real-world tasks.

\paragraph{Post-Hoc Explanation.}
The post-hoc methods do not require specific additional efforts during training and explainations are computed after training is finished. Most of these approaches invetsigate the gradient saliency in a trained neural network. For example, ~\citet{sundararajan2017axiomatic,smilkov2017smoothgrad,Bao2018DerivingMA} studied the integrated gradients of a model for interpretability.





\paragraph{Model-Based Approach.}
Another line of research focuses on developing models that self-explain the results. For example, module networks \cite{Andreas2016LearningTC} learn structures in addition to weights so that the learned structure can be used to interpret how the model processes and reasons over natural language. \citet{Johnson2017InferringAE} adapted the concept of module networks to the vision domain.

\subsection{Knowledge Distillation}

\citet{hinton2015distilling} proposed the teacher-student framework of knowledge distillation that transfers knowledge from a teacher model to a student model by optimizing the cross entropy loss. The most typical use case of knowledge distillation is model compression. A small model is distilled from a large pretrained model to achieve a more desirable complexity-effectiveness trade off \cite{Sanh2019DistilBERTAD,Jiao2019TinyBERTDB}. Knowledge distillation is also known to improve performance when the student model is of comparable size with the teacher model \cite{yim2017gift,furlanello2018born,wang2020knowledge} because it provides soft, continuous labels for more effective training. \cite{yoon2018invase} used knowledge distillation in selecting instance-wise features which is similar with rationalization.

\subsection{Learning Domain-Invariant Representations}

There are two main categories for learning domain-invariant representations---adversarial training and distribution matching. 
Adversarial training introduces an adversarial game where a discriminator learns to distinguish features extracted by an encoder \cite{Ganin2015UnsupervisedDA}. Distribution matching, on the other hand, is based on minimizing the distance between distributions in various forms \cite{Zellinger2017CentralMD,Gretton2006AKM,Li2015GenerativeMM}. 

\section{Distribution Matching for Rationalization}

In this section, we first introduce the standard rationalization framework \cite{lei2016rationalizing} and discuss our proposed method. Then we consider a more advanced rationalization variant (i.e., the game-theoretic approach introduced by \citet{chang2019game}) and discuss how to implement our framework on it.

\subsection{Preliminaries: The Rationalization Framework}

The input to the rationalization framework \cite{lei2016rationalizing} is a text sequence $\mathbf{x}=(x_1, x_2, \cdots, ,x_l)$ of length $l$, where each $x_i \in \mathcal{V}$ denotes the $i$-th token and $\mathcal{V}$ is the vocabulary. A generator $g$ is applied on $\mathbf{x}$ to obtain the rationale mask $\mathbf{z}$, i.e.,
\[
\mathbf{z} = g(\mathbf{x})
\]
where the rationale mask $\mathbf{z}$ is represented as a sequence of binary variables $\mathbf{z} = (z_1, z_2, \cdots, z_l)$. Each entry $z_i = 1$ means $x_i$ is selected as part of the rationales and $z_i = 0$ denotes the opposite. In other words, given the input text $\mathbf{x}$ and the mask $\mathbf{z}$, the rationale can be obtained as $\{x_i | z_i = 1\}$.

A classifier $c$ is applied on top of the rationale to obtain the model output distribution $\hat{p}(Y)$. Computationally, we use a lookup table to obtain the input embeddings $\mathbf{e}(\mathbf{x})$ and feed the masked embeddings to the classifier $c$ as follows:
\[
\hat{p}(Y) = c(\mathbf{z} \odot \mathbf{e}(\mathbf{x}))
\]
where $\odot$ denotes element-wise multiplication and $\hat{p}(Y)$ is a probability distribution over the classes.

Let $y$ be the groundtruth label in the label space $\mathcal{Y}$. The classification loss is written as a standard cross entropy loss:
\[
l_\text{cls} = - \log \hat{p}(Y = y)
\]

Additionally, it is desirable to control the sparsity and compactness of the rationales. To achieve this goal, the following regularization is applied:
\[
\Omega(\mathbf{z}) = \lambda_1 \|\mathbf{z}\|_1 + \lambda_2 \sum_{i = 2}^{l}|z_i - z_{i - 1}|
\]
with $\lambda_1$ and $\lambda_2$ being the coefficients.

The generator and the discriminator are jointly trained to minimize the overall loss function
\[
\min_{g, c} l_\text{cls} + \Omega(\mathbf{z}).
\]
Since $\mathbf{z}$ is discrete, the loss function is not differentiable w.r.t. the generator parameters. Methods like straight-through \cite{Bengio2013EstimatingOP} can be used for optimization.


\subsection{Distribution Matching for Rationalization (DMR)}

\begin{figure*}
    \centering
    \includegraphics[width=0.97\textwidth ]{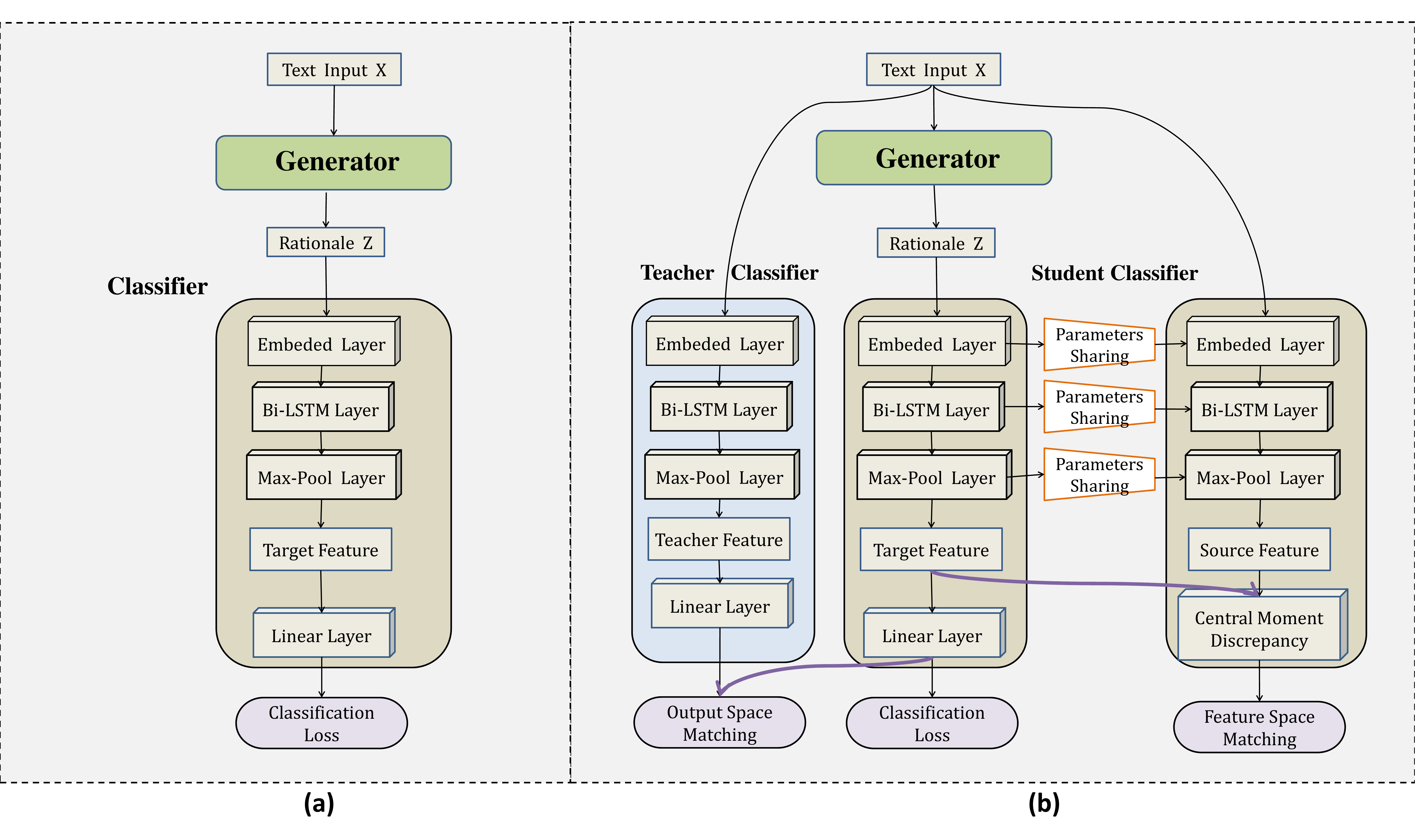}
    \caption{Comparison of (a) the baseline RNP framework, and (b) our proposed DMR framework. We run the classifier on the full input text for feature matching, and also train a student classifier for output matching.}
    \label{fig:DMR_frame}
\end{figure*}

Now we introduce our DMR method to improve rationalization based on distribution matching. Figure~\ref{fig:DMR_frame} illustrates the DMR model in comparison with the baseline RNP~\cite{lei2016rationalizing}. 

The underlying assumption and desideratum of our distribution matching approach is that the rationales and the original full input text should have similar feature and output distributions when the classifier is applied. Intuitively, interpreting model predictions is to explain how the classifier processes input information. Since rationales are to interpret and justify the predictions, when the input contains only the rationales, the classifier should learn features and make predictions in a very similar way compared to using the full original text as input. Based on this intuition, we propose to encourage similar distributions in both the feature and output spaces between the rationales and the full text input.

\subsubsection{Feature Space Matching}

The classifier $c$ can be instantiated as different models such recurrent neural networks \cite{Hochreiter1997LongSM}, convolutional networks \cite{Waibel1989PhonemeRU} and Transformers \cite{Vaswani2017AttentionIA}. Let $f$ be the function in classifier $c$ that maps word embeddings to network output features---e.g., the output of a max-pooling layer. Given a text sample $\mathbf{x}_i$ with rationale $\mathbf{z}_i$, it follows that $f(\mathbf{e}(\mathbf{x}_i))$ and $f(\mathbf{z}_i \odot \mathbf{e}(\mathbf{x}_i))$ are the features of the full input text and the rationales respectively. Denote these two features as $\mathbf{f}_i^x$ and $\mathbf{f}_i^z$ respectively.

Given a batch of $N$ training samples $\{\mathbf{x}_1, \mathbf{x}_2, \cdots, \mathbf{x}_N\}$ with rationales $\{\mathbf{z}_1, \mathbf{z}_2, \cdots, \mathbf{z}_N\}$ computed by the generator $g$, we employ the classifier $c$ to compute features $\{\mathbf{f}_1^x, \mathbf{f}_2^x, \cdots, \mathbf{f}_N^x\}$ and $\{\mathbf{f}_1^z, \mathbf{f}_2^z, \cdots, \mathbf{f}_N^z\}$. A central moment discrepancy regularizer \cite{Zellinger2017CentralMD} is employed to match the distributions in the feature space:
\[
l_\text{fm} = \|\mathbf{E}_x - \mathbf{E}_z\|_2 + \sum_{k = 2}^K \|\mathbf{C}_k^x - \mathbf{C}_k^z\|_2
\]
with
\[
\mathbf{E}_x = \frac{1}{N} \sum_{i = 1}^N \mathbf{f}_i^x
\]
\[
\mathbf{C}_k^x = \frac{1}{N} \sum_{i = 1}^N (\mathbf{f}_i^x - \mathbf{E}_x)^k
\]
where $\mathbf{E}_x$ the is empirical expectation of the features, and $\mathbf{C}_k^x$ is $k$-th order central moments of the feature coordinates. $\mathbf{E}_z$ and $\mathbf{C}_k^z$ are defined in a similar way. In practice, we compute the central moments up to the fifth order, i.e., $K = 5$. It is assumed that the features are distributed in the interval $[0, 1]$---e.g., the output of the sigmoid function; in other cases, constants might be added before each term \cite{Zellinger2017CentralMD}.

The feature space matching loss $l_\text{fm}$ enforces the rationales and the full input to have similar feature distributions when the classifier $c$ is applied.

\subsubsection{Output Space Matching}

A straightforward method to add training signals in the output space is to use a normal classification loss as in \cite{lei2016rationalizing}. However, following \cite{hinton2015distilling}, we believe only a categorical label is not sufficient to provide useful training signals. Our goal is to ensure that the rationales and the full input have similar output distributions, so knowledge distillation \cite{hinton2015distilling} is used for distribution matching in the probability output space.

Specifically, we first pretrain a teacher classifier $c_t$ that maps the full input text to the probability space using a standard classification loss. Let $\hat{p}_t(Y) = c_t(\mathbf{e}(\mathbf{x}))$ be the teacher model distribution of sample $\mathbf{x}$. The output space matching loss is written as the cross entropy between the teacher distribution $\hat{p}_t(Y)$ and the student distribution $\hat{p}(Y)$:
\[
l_\text{om} = \sum_{y = 1}^{|\mathcal{Y}|} - \hat{p}_t(Y = y) \log \hat{p}(Y = y).
\]

\subsubsection{Overall Loss Function}
\label{ssub:overall}
The overall loss function is formulated as the weighted sum of the feature and output space matching losses, along with the normal rationalization losses, i.e.,
\[
\min_c l_\text{cls} + \lambda_3 l_\text{fm} + \lambda_4 l_\text{om}
\]
\[
\min_g l_\text{cls} + \Omega(\mathbf{z})
\]
where $\lambda_3$ and $\lambda_4$ are the coefficients of the loss terms. Note that we apply the matching losses only on the classifier and do not backpropagate the gradients of $l_\text{fm}$ and $l_\text{om}$ to the generator $g$. Also the regularizer $\Omega(\mathbf{z})$ only depends on the generator. Although gradients from discriminator $d$ are not directly passed to $g$, the generators essentially benefit from the losses. 
Since we measure the accuracy of rationale prediction, the results improve if and and only if the generators improve. Therefore, the generators benefit a lot from our two regularization terms (though in an indirect manner).




\subsection{Extensions and Implementation}
Our above derivation is based on the original rationalization framework \cite{lei2016rationalizing}. However, our approach is general and applicable to different backbone methods. In our preliminary study, we experimented with multiple variants that improve over the original method and found that the CAR framework \cite{chang2019game} works particularly well. The CAR framework feed ground-truth label to the generator and proposed the cooperative and adversarial game mechanism.

We adopt a two-stage training scheme in our implementation. In the first stage, the teacher classifier $c_t$ is pretrained on the full dataset. After pretraining the teacher classifier $c_t$, we jointly train the student classifiers and the generators using the aforementioned losses.

\begin{table*}[]
    \centering
\vspace{0.5cm}
\small
\begin{tabular}{c|cccc|cccc|cccc}
\hline
\multicolumn{1}{c|}{\multirow{2}{*}{Beer}} & \multicolumn{4}{c|}{Appearance}                               & \multicolumn{4}{c|}{Aroma}                                    & \multicolumn{4}{c}{Palate}                                   \\
\multicolumn{1}{c|}{}                      & S    & P             & R             & \multicolumn{1}{c|}{F} & S    & P             & R             & \multicolumn{1}{c|}{F} & S    & P             & R             & \multicolumn{1}{c}{F} \\ \hline
CAR                                        & 11.9 & 76.2          & 49.3          & 59.9                   & 10.3 & 50.3          & 33.3          & 40.1                   & 10.2 & \textbf{56.6} & 46.2          & 50.9                   \\
DMR (ours)                                        & 11.7 & \textbf{83.6} & \textbf{52.8} & \textbf{64.7}          & 11.7 & \textbf{63.1}          & \textbf{47.6}          & \textbf{54.3}                   & 10.7 & 55.8          & \textbf{48.1} & \textbf{51.7}          \\
\hline
\multirow{2}{*}{Hotel} & \multicolumn{4}{c|}{Location}                        & \multicolumn{4}{c|}{Service}                         & \multicolumn{4}{c}{Cleanliness}                      \\
                      & S    & P             & R             & F             & S    & P             & R             & F             & S    & P             & R             & F             \\ \hline
CAR                   & 10.6 & 46.6          & 58.1          & 51.7          & 11.7 & 40.7          & 41.4          & 41.1          & 10.3 & 29.0          & 33.8          & 31.2          \\
DMR (ours)                   & 10.7 & \textbf{47.5} & \textbf{60.1} & \textbf{53.1} & 11.6 & \textbf{43.0} & \textbf{43.6} & \textbf{43.3} & 10.3 & \textbf{31.4} & \textbf{36.4} & \textbf{33.7} \\\hline
\end{tabular}

\caption{\normalsize Comparison with \emph{CAR} on both the beer review dataset and the hotel review dataset. S, P, R, and F1 represent the sparsity level, precision, recall, and F1 score respectively. We use the same (or similar) sparsity levels as previous work for fair comparison. All the baseline results are taken from \cite{chang2019game}.}
    \label{tab:table_game}
\end{table*}

\begin{table*}[!ht]
    \centering
\small
    \begin{tabular}{c|cccc|cccc|cccc}
\hline
\multicolumn{1}{c|}{\multirow{2}{*}{Beer}} & \multicolumn{4}{c|}{Appearance}                               & \multicolumn{4}{c|}{Aroma}                                    & \multicolumn{4}{c}{Palate}                                   \\
\multicolumn{1}{c|}{}                      & S    & P             & R             & \multicolumn{1}{c|}{F} & S    & P             & R             & \multicolumn{1}{c|}{F} & S    & P             & R             & \multicolumn{1}{c}{F} \\ \hline
RNP                                        & 7.9  & 13.5          & 5.8           & 8.1                    & 8.4  & 30.3          & 15.3          & 20.3                   & 9.1  & 28.2          & 17.2          & 21.4                   \\
3PLAYER                                    & 7.9  & 15.8          & 6.8           & 9.5                    & 8.4  & 48.9          & 24.4          & 32.6                   & 9.1  & 14.2          & 8.5           & 10.7                   \\
INVRAT                                     & 7.9  & 49.5          & 20.9          & 29.3                   & 8.4  & 48.2          & 24.4          & 32.4                   & 9.1  & 32.8          & 20.0          & 24.9                   \\

DMR (ours)                                        & 7.9  & \textbf{80.1} & \textbf{34.7} & \textbf{48.6}          & 8.9  & \textbf{50.3}          & \textbf{28.9} & \textbf{36.7}          & 9.6  & \textbf{49.7} & \textbf{38.2} & \textbf{43.2}          \\ \hline
RNP                                        & 15.8 & 13.5          & 11.3          & 12.3                   & 16.8 & 34.3          & 34.2          & 34.3                   & 18.1 & 19.8          & 23.8          & 21.6                   \\
3PLAYER                                    & 15.8 & 15.6          & 13.5          & 14.5                   & 16.8 & 35.7          & 35.9          & 35.8                   & 18.1 & 20.7          & 24.9          & 22.6                   \\
INVRAT                                     & 15.8 & 58.0          & 49.6          & 53.5                   & 16.8 & 42.7          & 42.5          & 42.6                   & 18.1 & \textbf{44.0} & \textbf{52.8}          & \textbf{48.0}                   \\

DMR (ours)                                        & 15.7 & \textbf{61.5} & \textbf{52.0} & \textbf{56.4}          & 16.8 & \textbf{47.6} & \textbf{51.3} & \textbf{49.4}          & 15.7 & 39.4          & 49.7 & 44.0          \\ \hline
RNP                                        & 23.7 & 26.3          & 33.1          & 29.3                   & 25.2 & 40.0          & 60.1          & 48.0                   & 27.2 & 19.2          & 33.8          & 24.5                   \\
3PLAYER                                    & 23.7 & 12.6          & 15.9          & 14.0                   & 25.2 & 33.0          & 49.7          & 39.7                   & 27.2 & 22.0          & 39.3          & 28.2                   \\
    INVRAT                                     & 23.7 & 54.0          & \textbf{69.2}          & 60.7                   & 25.2 & 44.7 & 67.4          & 53.8                   & 27.2 & 26.5          & 46.9          & 33.9                   \\

DMR (ours)                                        & 21.2 & \textbf{58.9} & 67.4 & \textbf{62.9}          & 25.0 & \textbf{44.7}          & \textbf{71.8} & \textbf{55.1}          & 27.2 & \textbf{28.0} & \textbf{61.3} & \textbf{38.4}          \\ \hline
\end{tabular}
    \caption{\normalsize Comparsion with \emph{RNP}, \emph{3PLAYER} and \emph{INVRAT} on the beer review dataset. S, P, R, and F1 represent the sparsity level, precision, recall, and F1 score respectively. We use the same (or similar) sparsity levels as previous work for fair comparison. All the baseline results are taken from \cite{chang2020invariant}.}
    \label{tab:table_invariant}
\end{table*}

\begin{table*}[!t]
\centering
\small
\begin{tabular}{c|ccc|ccc|ccc}
\hline
\multicolumn{1}{l|}{} & \multicolumn{3}{c}{appearance} & \multicolumn{3}{c|}{aroma}     & \multicolumn{3}{c}{palate}                        \\
input                 & precision & recall  & f1-score & precision & recall  & f1-score & precision & recall  & f1-score                     \\ \hline
rationales            & 92.09   & 88.56 & 90.29  & 93.47   & 79.23 & 85.77  & 93.87   & 71.79 & \multicolumn{1}{c}{81.36} \\
full texts            & 91.54   & 91.88 & 91.71  & 94.34   & 79.26 & 86.15  & 94.52   & 70.08 & \multicolumn{1}{c}{80.49} \\ \hline
\end{tabular}
\centering
\caption{\normalsize Comparison of classification performances using rationales and full texts}
    \label{tab:rationales classification}
\end{table*}

\section{Experiments}
\subsection{Datasets}
To evaluate the performance of our DMR framework, we use the multi-aspect beer and hotel datasets, which are commonly used in the field of rationalization.

\paragraph{Beer reviews:} The beer review dataset~\cite{McAuley2012LearningAA} is a multi-aspect sentiment classification dataset, where each review of a beer consists of a plain-text comment and ratings from three aspects including appearance, aroma and palate. 


\paragraph{Hotel reviews:} The hotel review dataset~\cite{Bao2018DerivingMA} is another multi-aspect  sentiment classification dataset. 
The dataset contains reviews of hotels from three aspects including location, cleanliness, and service. 
Each review also has a rating on a scale of 0-5 stars. 

We preprocess both datasets in the same setting as~\cite{chang2019game} for fair comparison. 


\subsection{Baselines}

We consider the following baselines for comparison in our experiments:
\begin{itemize}
    \item \textbf{RNP:} RNP is the original rationalization framework proposed in ~\cite{lei2016rationalizing}.
    The generator selects text segments as rationales, and the predictor is fed with rationales for label classification. 
    RNP aims to maximize the mutual information between
    rationales and labels and the rationales are constrained to be both sparse and continuous. 
    
    \item \textbf{3PLAYER:} The 3PLAYER method is an enhancement of RNP~\cite{yu2019rethinking}, which alleviates the degeneration problem of RNP by introducing an extra complement predictor. The complement predictor tries to maximize the predictive accuracy from unselected words and plays an adversarial game with generator. 
    
    \item \textbf{INVRAT:} INVRAT introduces a game-theoretic invariant criterion as the objective and aims to learn environment invariant representations~\cite{chang2020invariant}. 
    

    \item \textbf{CAR:} CAR proposes a game theoretic approach to class-wise selective rationalization~\cite{chang2019game}. The approach produces both positive and negative rationales. This is the direct baseline of our results because we use CAR as our backbone model.
    
\end{itemize}

To seek for fair comparisons, the generators, the predictors, and the discriminators of all the baselines and our method use the same architecture as in the previous work \cite{chang2019game}. In addition, the sparsity and continuity constraints follow the same form and all methods are adjusted to have a comparable level of sparsity in the experiments. 
In our experiment, we implement our DMR framework directly based on the CAR model. 

Following previous work \cite{chang2019game}, the hidden unit size and the embedding dimension of the teacher discriminator are set as 100, while those of the generators and the student discriminator are set as 102 for two extra class label dimensions. 

\begin{table*}[!t]
\centering
\vspace{0.5cm}
\small
\begin{tabular}{c|cccc|cccc|cccc}
\hline
\multicolumn{1}{c|}{\multirow{2}{*}{Beer}} & \multicolumn{4}{c|}{Appearance}                               & \multicolumn{4}{c|}{Aroma}                                    & \multicolumn{4}{c}{Palate}                           \\
\multicolumn{1}{c|}{}                      & S    & P             & R             & \multicolumn{1}{c|}{F} & S    & P             & R             & \multicolumn{1}{c|}{F} & S    & P             & R             & F             \\ \hline
DMR                                        & 8.8  & \textbf{78.4} & \textbf{37.3} & \textbf{50.6}          & 8.8  & 48.7          & \textbf{27.6} & \textbf{35.3}          & 8.9  & \textbf{60.2} & \textbf{43.2} & \textbf{50.3} \\ 
~~ - fm                                 & 8.9  & 74.5          & 35.7          & 48.3                   & 8.4  & 49.5          & 26.7          & 34.7                   & 8.9  & 58.6          & 42.2          & 49.1          \\
~~ - fm\&om                           & 8.8  & 70.0          & 33.4          & 45.2                   & 7.7  & \textbf{50.2} & 24.8          & 33.2                   & 8.9  & 54.3          & 38.9          & 45.3          \\ \hline
DMR                                        & 12.3 & \textbf{82.6} & \textbf{55.0} & \textbf{66.1}          & 13.0 & \textbf{61.8} & \textbf{51.5} & \textbf{56.1}          & 11.5 & \textbf{55.8} & \textbf{51.6} & \textbf{53.6} \\ 
~~ - fm                                 & 12.4 & 77.0          & 51.5          & 63.5                   & 12.0 & 60.4          & 46.5          & 52.6                   & 12.3 & 51.3          & 50.5          & 50.9          \\
~~ - fm\&om                           & 12.3 & 74.5          & 49.6          & 59.6                   & 12.9 & 55.2          & 45.6          & 50.0                   & 12.1 & 50.6          & 49.1          & 49.8          \\\hline
DMR                                        & 16.0 & \textbf{72.8} & \textbf{63.0} & \textbf{67.5}          & 16.2 & 52.3          & \textbf{54.6} & \textbf{53.4}          & 16.4 & \textbf{45.2} & \textbf{59.5} & \textbf{51.4} \\ 
~~ - fm                                 & 16.0 & 69.0          & 59.7          & 64.0                   & 15.5 & \textbf{52.4} & 52.2          & 52.3                   & 16.0 & 38.4          & 49.4          & 43.2          \\
~~ - fm\&om                           & 16.0 & 63.9          & 55.4          & 59.4                   & 16.1 & 36.2          & 37.5          & 36.8                   & 16.3 & 37.7          & 49.3          & 42.7          \\\hline
DMR                                        & 24.0 & \textbf{57.7} & \textbf{74.7} & \textbf{65.1}          & 24.1 & \textbf{46.1} & \textbf{71.3} & \textbf{56.0}          & 23.8 & \textbf{36.6} & \textbf{70.2} & \textbf{48.1} \\ 
~~ - fm                                 & 23.9 & 54.1          & 69.9          & 61.0                   & 24.0 & 44.0          & 67.8          & 53.4                   & 24.0 & 29.8          & 57.4          & 39.2          \\
~~ - fm\&om                           & 23.9 & 53.9          & 69.7          & 60.8                   & 24.1 & 43.0          & 66.6          & 52.2                   & 23.9 & 28.5          & 54.8          & 37.5          \\\hline
\end{tabular}
\caption{\normalsize Ablation study. ``- fm'' means removing the feature space matching loss, and ``- fm\&om'' means removing both feature and output space matching losses.}
\label{tab:ablation_study}
\end{table*}

\begin{table*}[]
    \centering
\small

\begin{tabular}{c|cccc|cccc|cccc}
\hline
\multirow{2}{*}{Beer} & \multicolumn{4}{c|}{Appearance}                        & \multicolumn{4}{c|}{Aroma}                         & \multicolumn{4}{c}{Palate}                      \\
                      & S    & P             & R             & F             & S    & P             & R             & F             & S    & P             & R             & F             \\ \hline
DMR$_{CORAL}$                   & 11.9 & 79.0          & 50.8          & 61.9          & 11.7 & 60.3          & 45.2          & 51.6          & 10.1 & 47.0          & 38.0          & 42.0          \\
DMR$_{MMD}$                   & 11.7 & 81.0          & 51.4          & 62.9          & 11.5 & 50.0          & 36.9          & 42.4          & 10.6 & 44.5          & 38.0          & 41.0          \\
DMR (ours)                   & 11.7 & \textbf{83.6} & \textbf{52.8} & \textbf{64.7} & 11.7 & \textbf{63.1} & \textbf{47.6} & \textbf{54.3} & 10.7 & \textbf{55.8} & \textbf{48.1} & \textbf{51.7} \\\hline
\end{tabular}
\caption{\normalsize Comparison of Different Matching Loss Selected.}
    \label{tab:table_loss_sensitivity}
\end{table*}

\begin{figure}
    \centering
    \begin{subfigure}{0.4\textwidth}
    \includegraphics[width=0.8\linewidth]{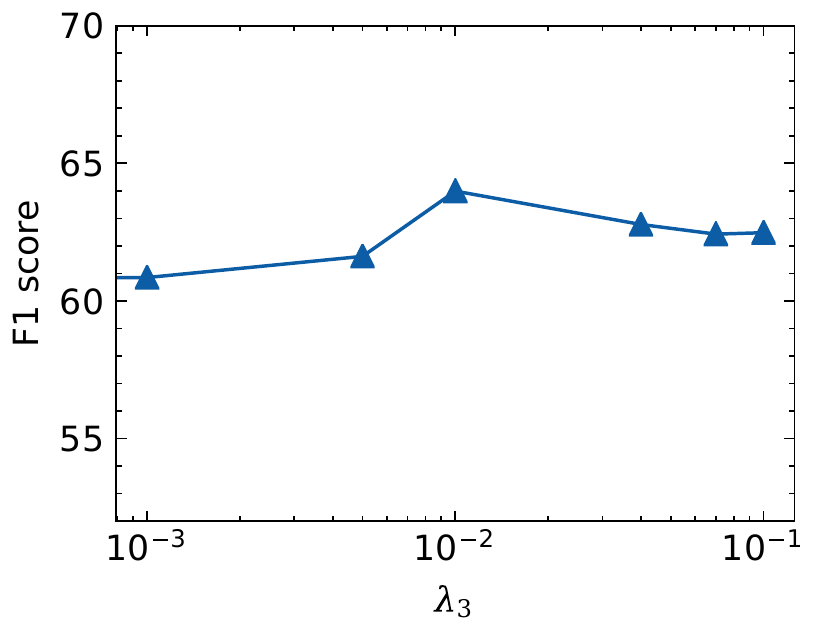}
    \caption{}
    \end{subfigure}
    \begin{subfigure}{0.4\textwidth}
    \includegraphics[width=0.8\linewidth]{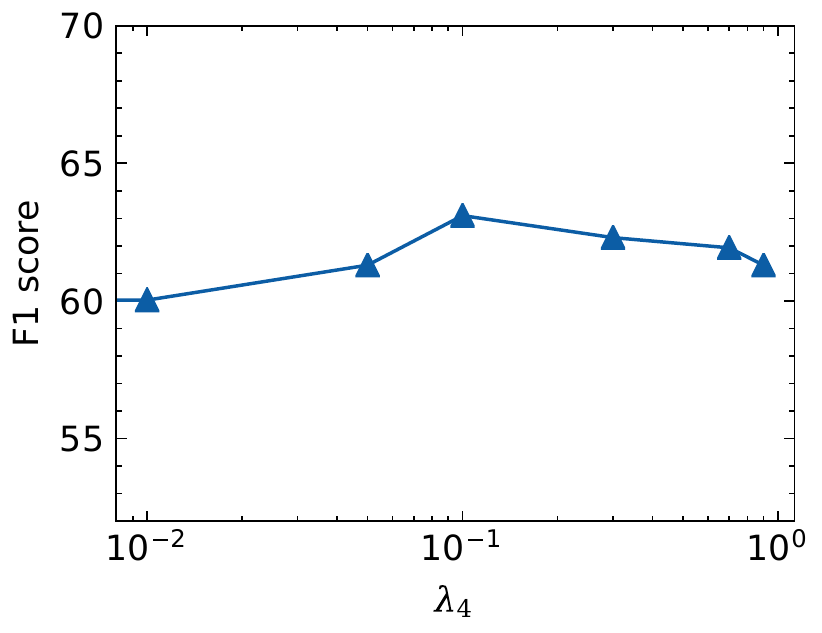}
    \caption{}
    \end{subfigure}
    \caption{Parameter Sensitivity of $\lambda_3$ and $\lambda_4$}
    \label{fig:lambdas_sensitivity}
\end{figure}

\subsection{Quantitative Evaluation}

In this section, we evaluate the performance of DMR and compare its performance against with the state of arts methods on the beer and the hotel review datasets.

We train our models using a balanced training dataset as in~\cite{chang2019game} and evaluate the performances on the test sets with human annotated rationales.
Since  rationales generated from CAR were based on ground truth label,  we also infer our rationales condition on true labels. As results shown in Table~\ref{tab:table_game}, DMR outperforms CAR in all aspects of two datasets.

To further compare with \emph{RNP}, \emph{INVRAT} and \emph{3PLAYER},  we adjust the sparsity levels to obtain the same rationale lengths on the beer dataset as in~\cite{chang2020invariant}. The data split and processing of the beer review dataset in ~\cite{chang2020invariant} are not available. 
For  fair comparsion,  we train and validate the models using the same beer review dataset but with different data split and processing, and the performances of rationales are evaluated on the same annotated test set.
As shown in Table \ref{tab:table_invariant}, DMR obtains the best performance in almost all metrics and sparsity combinations. 
In addition, our DMR method does not need ground-truth labels to generate rationales, as the predicted labels provided by the teacher classifier are used instead.

The classification results in the Table~\ref{tab:rationales classification} show that better rationalization does not substantially improve classification, but with our method, training a classifier on the rationales is able to achieve performance comparable to using full text. 

The results in Table~\ref{tab:table_game} and ~\ref{tab:table_invariant} reveal the effectiveness of distribution matching.  Compared with existing approaches, DMR is able to extract more accurate rationales and the advantages can be extended to all sparsity levels. And Table~\ref{tab:rationales classification} presents that extracted rationales by our DMR are comparable to full texts on the classification.

\subsection{Ablation Study}
\subsubsection{Effectiveness of our Matching Losses}
We conducted ablation studies to understand the importance of feature space matching and output space matching in the training process.

In Table~\ref{tab:ablation_study}, we show the performances of DMR with different matching losses removed (row begin with ``-'') under different levels of sparsity. The rows with ``-fm'' stand for models that have the feature space matching loss removed, and the rows with ``-fm\&om'' correspond to models trained without neither the feature space matching nor the output space matching. 

As shown in the table, both feature space matching and output space matching contribute to the performance of our method. The improvements brought by both methods are consistent and substantial across different settings, which validates our motivation in the previous sections.

\begin{table*}[!ht]
\resizebox{\textwidth}{!}{%
\small
\begin{tabular}{@{}c|cc@{}}
\hline

Aspect     & DMR Rationale       & CAR Rationale\\ \hline

Appearance 
& \begin{tabular}[c]{@{}p{9cm}@{}}
\\
\textbf{\uline{tangerine pour with a small white head that clings to the edge of the glass .}}
the hopping is smooth and mild , but the bitterness does gradually build , although it reminded me more of an english bitter instead of an american ipa . the malts come out as fruity with some honey . medium-light body with decent carbonation . i ca n't give it a glowing review because its not a great beer . pyramid seems to be very hit and miss , and this is a miss . 
\end{tabular} & 

\begin{tabular}[c]{@{}p{9cm}@{}}
\\
\uline{\textbf{tangerine pour with a} small white head that clings to the edge of the glass .}
the hopping is smooth and mild , but the bitterness does gradually build , although it reminded me more of an english bitter instead of an american \textbf{ipa . the} malts come out as fruity with some honey . medium-light body with decent carbonation . i ca n't give it a glowing review because its not \textbf{a great beer . pyramid seems} to be very hit and miss , and this is a miss . 
 \end{tabular} \\  \hline

Aroma      
& \begin{tabular}[c]{@{}p{9cm}@{}}
\\
appearence: pours a crystal clear amber with a thin , bubbly white head that dies to a collar. \textbf{\uline{smell : solid belgian pale ale malt and hop characteristics throughout, with that perfect yeast tinge.}} \textbf{taste} and mouthfeel : rich , full , thirst-quenching , and  smooth . very balanced and tasty , with the perfect mouthfeel . 

\\ 
\end{tabular}                                                                                                          

& \begin{tabular}[c]{@{}p{9cm}@{}}
\\
appearence : pours a crystal clear amber with a thin , bubbly white head that dies to a collar . \uline{smell : solid belgian pale ale malt and hop characteristics throughout , with that perfect yeasttinge}. taste \textbf{and mouthfeel : rich , full , thirst-quenching , and  smooth . very balanced and tasty , with the} perfect mouthfeel . 
\\  
\end{tabular} \\  \hline

Palate     
& \begin{tabular}[c]{@{}p{9cm}@{}}
pours an amber with an orange hue . two inch white head fades quickly . very little 
lacing . smells like bread , not much else . taste some sweet malt , and grass . not 
much better than a macro . \textbf{\uline{lighter body with lots of carbonation.}} \textbf{not} a lot of flavor but this is a refreshing beer . i have no problem drinking these , i just would n't pursue it .
\end{tabular}                                                                            
& \begin{tabular}[c]{@{}p{9cm}@{}}
\\
\textbf{pours an} amber with an orange hue . two inch white head fades quickly . very little 
lacing . smells like bread , not much else . taste some sweet malt , and grass . not 
much better than a macro . \textbf{\uline{lighter body} \uline{with lots of carbonation.}} not a lot of flavor but \textbf{this is a } beer. i have no problem drinking these , i just would n't pursue it .
\end{tabular}                    \\ \hline
\end{tabular}%
}
\caption{\normalsize Examples of rationales generated by our DMR method and the baseline CAR method on the three aspects of the beer dataset. Underlined words are the human annotated labels, and bold word are predicted positive rationales.}
\label{tab:Rationale_Case}
\end{table*}

\begin{table*}[!ht]
\resizebox{\textwidth}{!}{%
\small
\begin{tabular}{@{}c|cc@{}}
\hline

Aspect     & DMR Rationale       \\ \hline

Appearance 
& \begin{tabular}[c]{@{}p{18cm}@{}}
\\
 \textbf{\uline{the beige head is comprised of }}\uline{medium-sized bubbles and}\textbf{\uline{ slowly , but inevitably ,}}\uline{ recedes to a thin strip ; there is some lacing adhering to the glass .} black malt ( i think ) lends the beer a not pleasantly bitter and toasted flavour . 
 \\
\end{tabular} & 

\\  \hline

Aroma      
& \begin{tabular}[c]{@{}p{18cm}@{}}

\\
pours with a nice foamy frothy off white head that lasts and a little lace . color is an ever so slightly \textbf{hazy amber . \uline{aroma is malty , grassy , hoppy , and bready beer .} flavor 's very similar along with} pretzels and with bitterness coming out at the end.
\end{tabular}                                                                                            \\  \hline

Palate     
& \begin{tabular}[c]{@{}p{18cm}@{}}
\\
typical of a hefeweizen taste shows notes of orange and the typical hefeweizen taste ( cloves and bananas ) . \textbf{\uline{smooth and very effervescent . almost no}} \uline{bitterness too . }\textbf{very drinkable and refreshing .} a nice \textbf{hefeweizen} . 
\end{tabular}                                                           
\\ \hline
\end{tabular}%
}
\caption{\normalsize Failure examples of rationales generated by our DMR method on the beer dataset. The underlined words are the human annotated labels, and the bold words are predicted positive rationales.}
\label{tab:failure_example}
\end{table*}

\subsubsection{Comparison of different Feature Space Matching Losses}
Many studies have shown that CMD outperforms MMD~\cite{Li2015GenerativeMM} and CORAL~\cite{sun2016deep} for its efficiency and invariance to different styles of input. 
In our experiments, we also find that using CMD matching can provide more stable and generally better results. 
Results in Table~\ref{tab:table_loss_sensitivity} show that using CMD can always lead to the best results with respect to the F1 scores on all aspects of beer review dataset.
\subsubsection{Sensitivities of Hyper-parameters}
We also studies the influences of different values of hyper-parameters $\lambda_3$ and $\lambda_4$ in section~\ref{ssub:overall}, respectively. 
As presented in Figure~\ref{fig:lambdas_sensitivity}, the overall performance of our method is not sensitive to either the values of $\lambda_3$ or $\lambda_4$. 
\subsection{Case Studies}
In this section, we visualize the rationales generated by our DMR framework and the CAR framework.
As presented in Table~\ref{tab:Rationale_Case}, rationales generated by DMR are more accurate. For example, the DMR framework selects exactly the same rationales as the human annotations, while the CAR framework only finds a few related words combined with many unrelated words, specially in appearance and aroma aspects, which shows that DMR can extract meaningful and accurate rationales.In addition, we also show some failure examples in Table~\ref{tab:failure_example}.
\section{Conclusions}

In this paper, we propose a novel rationalization framework based on distribution matching called DMR. DMR aims to match rationales and the input text in both the feature space and the output space. 
For feature space matching, we formulate it as minimization of the central moment discrepancy (CMD) between input text features and the rationale features. For the output space matching, we transfer the knowledge from the output distribution of the original full text to that of the rationales in a teacher-student distillation framework.
The framework is highly flexible and can be applied to many existing rationale extraction methods. Extensive experiments show that the DMR framework outperforms state-of-the-art methods in most experimental settings.
Ablation studies show that both feature space matching and output space matching contribute to the final performance.
Moreover, case analysis show that DMR provides more meaningful and accurate rationales. In the future, it will be intriguing to apply our method to more interpretability settings such as non-classification tasks.

\section{Acknowledgments}
This work is funded by National Key R\&D Program of China (2020AAA0105200) and supported by Beijing Academy of Artificial Intelligence (BAAI).

\bibliography{aaai2021}

\begin{thebibliography}{34}
\providecommand{\natexlab}[1]{#1}
\providecommand{\url}[1]{\texttt{#1}}
\providecommand{\urlprefix}{URL }
\expandafter\ifx\csname urlstyle\endcsname\relax
  \providecommand{\doi}[1]{doi:\discretionary{}{}{}#1}\else
  \providecommand{\doi}{doi:\discretionary{}{}{}\begingroup
  \urlstyle{rm}\Url}\fi

\bibitem[{Andreas et~al.(2016)Andreas, Rohrbach, Darrell, and
  Klein}]{Andreas2016LearningTC}
Andreas, J.; Rohrbach, M.; Darrell, T.; and Klein, D. 2016.
\newblock Learning to Compose Neural Networks for Question Answering.
\newblock \emph{ArXiv} abs/1601.01705.

\bibitem[{Bao et~al.(2018)Bao, Chang, Yu, and Barzilay}]{Bao2018DerivingMA}
Bao, Y.; Chang, S.; Yu, M.; and Barzilay, R. 2018.
\newblock Deriving Machine Attention from Human Rationales.
\newblock In \emph{EMNLP}.

\bibitem[{Bastings, Aziz, and Titov(2019)}]{Bastings2019InterpretableNP}
Bastings, J.; Aziz, W.; and Titov, I. 2019.
\newblock Interpretable Neural Predictions with Differentiable Binary
  Variables.
\newblock In \emph{ACL}.

\bibitem[{Bengio, L{\'e}onard, and Courville(2013)}]{Bengio2013EstimatingOP}
Bengio, Y.; L{\'e}onard, N.; and Courville, A.~C. 2013.
\newblock Estimating or Propagating Gradients Through Stochastic Neurons for
  Conditional Computation.
\newblock \emph{ArXiv} abs/1308.3432.

\bibitem[{Chang et~al.(2019)Chang, Zhang, Yu, and Jaakkola}]{chang2019game}
Chang, S.; Zhang, Y.; Yu, M.; and Jaakkola, T. 2019.
\newblock A Game Theoretic Approach to Class-wise Selective Rationalization.
\newblock In \emph{NeurIPS}, 10055--10065.

\bibitem[{Chang et~al.(2020)Chang, Zhang, Yu, and
  Jaakkola}]{chang2020invariant}
Chang, S.; Zhang, Y.; Yu, M.; and Jaakkola, T.~S. 2020.
\newblock Invariant rationalization.
\newblock \emph{arXiv preprint arXiv:2003.09772} .

\bibitem[{Chen et~al.(2018)Chen, Song, Wainwright, and
  Jordan}]{Chen2018LearningTE}
Chen, J.; Song, L.; Wainwright, M.~J.; and Jordan, M.~I. 2018.
\newblock Learning to Explain: An Information-Theoretic Perspective on Model
  Interpretation.
\newblock In \emph{ICML}.

\bibitem[{Devlin et~al.(2019)Devlin, Chang, Lee, and
  Toutanova}]{Devlin2019BERTPO}
Devlin, J.; Chang, M.-W.; Lee, K.; and Toutanova, K. 2019.
\newblock BERT: Pre-training of Deep Bidirectional Transformers for Language
  Understanding.
\newblock \emph{ArXiv} abs/1810.04805.

\bibitem[{DeYoung et~al.(2019)DeYoung, Jain, Rajani, Lehman, Xiong, Socher, and
  Wallace}]{deyoung2019eraser}
DeYoung, J.; Jain, S.; Rajani, N.~F.; Lehman, E.; Xiong, C.; Socher, R.; and
  Wallace, B.~C. 2019.
\newblock ERASER: A Benchmark to Evaluate Rationalized NLP Models.
\newblock \emph{arXiv preprint arXiv:1911.03429} .

\bibitem[{Furlanello et~al.(2018)Furlanello, Lipton, Tschannen, Itti, and
  Anandkumar}]{furlanello2018born}
Furlanello, T.; Lipton, Z.~C.; Tschannen, M.; Itti, L.; and Anandkumar, A.
  2018.
\newblock Born again neural networks.
\newblock \emph{arXiv preprint arXiv:1805.04770} .

\bibitem[{Ganin and Lempitsky(2015)}]{Ganin2015UnsupervisedDA}
Ganin, Y.; and Lempitsky, V.~S. 2015.
\newblock Unsupervised Domain Adaptation by Backpropagation.
\newblock \emph{ArXiv} abs/1409.7495.

\bibitem[{Gretton et~al.(2006)Gretton, Borgwardt, Rasch, Sch{\"o}lkopf, and
  Smola}]{Gretton2006AKM}
Gretton, A.; Borgwardt, K.~M.; Rasch, M.~J.; Sch{\"o}lkopf, B.; and Smola,
  A.~J. 2006.
\newblock A Kernel Method for the Two-Sample-Problem.
\newblock In \emph{NeurIPS}.

\bibitem[{Hinton, Vinyals, and Dean(2015)}]{hinton2015distilling}
Hinton, G.; Vinyals, O.; and Dean, J. 2015.
\newblock Distilling the knowledge in a neural network.
\newblock \emph{arXiv preprint arXiv:1503.02531} .

\bibitem[{Hochreiter and Schmidhuber(1997)}]{Hochreiter1997LongSM}
Hochreiter, S.; and Schmidhuber, J. 1997.
\newblock Long Short-Term Memory.
\newblock \emph{Neural Computation} 9: 1735--1780.

\bibitem[{Jiao et~al.(2019)Jiao, Yin, Shang, Jiang, Chen, Li, Wang, and
  Liu}]{Jiao2019TinyBERTDB}
Jiao, X.; Yin, Y.; Shang, L.; Jiang, X.; Chen, X.; Li, L.; Wang, F.; and Liu,
  Q. 2019.
\newblock TinyBERT: Distilling BERT for Natural Language Understanding.
\newblock \emph{ArXiv} abs/1909.10351.

\bibitem[{Johnson et~al.(2017)Johnson, Hariharan, van~der Maaten, Hoffman,
  Fei-Fei, Zitnick, and Girshick}]{Johnson2017InferringAE}
Johnson, J.; Hariharan, B.; van~der Maaten, L.; Hoffman, J.; Fei-Fei, L.;
  Zitnick, C.~L.; and Girshick, R.~B. 2017.
\newblock Inferring and Executing Programs for Visual Reasoning.
\newblock \emph{ICCV} 3008--3017.

\bibitem[{Lei, Barzilay, and Jaakkola(2016)}]{lei2016rationalizing}
Lei, T.; Barzilay, R.; and Jaakkola, T. 2016.
\newblock Rationalizing neural predictions.
\newblock \emph{arXiv preprint arXiv:1606.04155} .

\bibitem[{Li, Swersky, and Zemel(2015)}]{Li2015GenerativeMM}
Li, Y.; Swersky, K.; and Zemel, R.~S. 2015.
\newblock Generative Moment Matching Networks.
\newblock In \emph{ICML}.

\bibitem[{McAuley, Leskovec, and Jurafsky(2012)}]{McAuley2012LearningAA}
McAuley, J.~J.; Leskovec, J.; and Jurafsky, D. 2012.
\newblock Learning Attitudes and Attributes from Multi-aspect Reviews.
\newblock \emph{ICDM} 1020--1025.

\bibitem[{Paranjape et~al.(2020)Paranjape, Joshi, Thickstun, Hajishirzi, and
  Zettlemoyer}]{paranjape2020information}
Paranjape, B.; Joshi, M.; Thickstun, J.; Hajishirzi, H.; and Zettlemoyer, L.
  2020.
\newblock An Information Bottleneck Approach for Controlling Conciseness in
  Rationale Extraction.
\newblock \emph{arXiv preprint arXiv:2005.00652} .

\bibitem[{Peters et~al.(2018)Peters, Neumann, Iyyer, Gardner, Clark, Lee, and
  Zettlemoyer}]{DBLP:journals/corr/abs-1802-05365}
Peters, M.~E.; Neumann, M.; Iyyer, M.; Gardner, M.; Clark, C.; Lee, K.; and
  Zettlemoyer, L. 2018.
\newblock Deep contextualized word representations.
\newblock \emph{arXiv} abs/1802.05365.

\bibitem[{Radford(2018)}]{Radford2018ImprovingLU}
Radford, A. 2018.
\newblock Improving Language Understanding by Generative Pre-Training.

\bibitem[{Sanh et~al.(2019)Sanh, Debut, Chaumond, and
  Wolf}]{Sanh2019DistilBERTAD}
Sanh, V.; Debut, L.; Chaumond, J.; and Wolf, T. 2019.
\newblock DistilBERT, a distilled version of BERT: smaller, faster, cheaper and
  lighter.
\newblock \emph{ArXiv} abs/1910.01108.

\bibitem[{Smilkov et~al.(2017)Smilkov, Thorat, Kim, Vi{\'e}gas, and
  Wattenberg}]{smilkov2017smoothgrad}
Smilkov, D.; Thorat, N.; Kim, B.; Vi{\'e}gas, F.; and Wattenberg, M. 2017.
\newblock Smoothgrad: removing noise by adding noise.
\newblock \emph{arXiv preprint arXiv:1706.03825} .

\bibitem[{Sun and Saenko(2016)}]{sun2016deep}
Sun, B.; and Saenko, K. 2016.
\newblock Deep coral: Correlation alignment for deep domain adaptation.
\newblock In \emph{ECCV}, 443--450. Springer.

\bibitem[{Sundararajan, Taly, and Yan(2017)}]{sundararajan2017axiomatic}
Sundararajan, M.; Taly, A.; and Yan, Q. 2017.
\newblock Axiomatic attribution for deep networks.
\newblock In \emph{Proceedings of the 34th International Conference on Machine
  Learning-Volume 70}, 3319--3328. JMLR. org.

\bibitem[{Vaswani et~al.(2017)Vaswani, Shazeer, Parmar, Uszkoreit, Jones,
  Gomez, Kaiser, and Polosukhin}]{Vaswani2017AttentionIA}
Vaswani, A.; Shazeer, N.; Parmar, N.; Uszkoreit, J.; Jones, L.; Gomez, A.~N.;
  Kaiser, L.; and Polosukhin, I. 2017.
\newblock Attention is All you Need.
\newblock \emph{ArXiv} abs/1706.03762.

\bibitem[{Waibel et~al.(1989)Waibel, Hanazawa, Hinton, Shikano, and
  Lang}]{Waibel1989PhonemeRU}
Waibel, A.~H.; Hanazawa, T.; Hinton, G.~E.; Shikano, K.; and Lang, K.~J. 1989.
\newblock Phoneme recognition using time-delay neural networks.
\newblock \emph{ICASSP} 37: 328--339.

\bibitem[{Wang and Yoon(2020)}]{wang2020knowledge}
Wang, L.; and Yoon, K.-J. 2020.
\newblock Knowledge distillation and student-teacher learning for visual
  intelligence: A review and new outlooks.
\newblock \emph{arXiv preprint arXiv:2004.05937} .

\bibitem[{Yang et~al.(2019)Yang, Dai, Yang, Carbonell, Salakhutdinov, and
  Le}]{Yang2019XLNetGA}
Yang, Z.; Dai, Z.; Yang, Y.; Carbonell, J.~G.; Salakhutdinov, R.; and Le, Q.~V.
  2019.
\newblock XLNet: Generalized Autoregressive Pretraining for Language
  Understanding.
\newblock In \emph{NeurIPS}.

\bibitem[{Yim et~al.(2017)Yim, Joo, Bae, and Kim}]{yim2017gift}
Yim, J.; Joo, D.; Bae, J.; and Kim, J. 2017.
\newblock A gift from knowledge distillation: Fast optimization, network
  minimization and transfer learning.
\newblock In \emph{CVPR}, 4133--4141.

\bibitem[{Yoon, Jordon, and van~der Schaar(2019)}]{yoon2018invase}
Yoon, J.; Jordon, J.; and van~der Schaar, M. 2019.
\newblock {INVASE}: Instance-wise Variable Selection using Neural Networks.
\newblock In \emph{ICLR}.
\newblock \urlprefix\url{https://openreview.net/forum?id=BJg_roAcK7}.

\bibitem[{Yu et~al.(2019)Yu, Chang, Zhang, and Jaakkola}]{yu2019rethinking}
Yu, M.; Chang, S.; Zhang, Y.; and Jaakkola, T.~S. 2019.
\newblock Rethinking cooperative rationalization: Introspective extraction and
  complement control.
\newblock \emph{arXiv preprint arXiv:1910.13294} .

\bibitem[{Zellinger et~al.(2017)Zellinger, Grubinger, Lughofer,
  Natschl{\"a}ger, and Saminger-Platz}]{Zellinger2017CentralMD}
Zellinger, W.; Grubinger, T.; Lughofer, E.; Natschl{\"a}ger, T.; and
  Saminger-Platz, S. 2017.
\newblock Central Moment Discrepancy (CMD) for Domain-Invariant Representation
  Learning.
\newblock \emph{ArXiv} abs/1702.08811.

\end{thebibliography}

\end{document}